# The IHS Transformations Based Image Fusion

Firouz Abdullah Al-Wassai[1], N.V. Kalyankar[2], Ali A. Al-Zuky[3]

*[1] Research Student, Computer Science Dept., Yeshwant College, (SRTMU), Nanded, India

fairozwaseai@yahoo.com [1]

[2] Principal, Yeshwant Mahavidyala College, Nanded, India

drkalyankarnv@rediffmail.com [2]

[3] Assistant Professor, Dept. of Physics, College of Science, Mustansiriyah Un. , Baghdad – Iraq.

dr.alialzuky@yahoo.com [3]

*Abstract:* The IHS sharpening technique is one of the most commonly used techniques for sharpening. Different transformations have been developed to transfer a color image from the RGB space to the IHS space. Through literature, it appears that, various scientists proposed alternative IHS transformations and many papers have reported good results whereas others show bad ones as will as not those obtained which the formula of IHS transformation were used. In addition to that, many papers show different formulas of transformation matrix such as IHS transformation. This leads to confusion what is the exact formula of the IHS transformation?. Therefore, the main purpose of this work is to explore different IHS transformation techniques and experiment it as IHS based image fusion. The image fusion performance was evaluated, in this study, using various methods to estimate the quality and degree of information improvement of a fused image quantitatively.

## I. INTRODUCTION

Remote sensing offers a wide variety of image data with different characteristics in terms of temporal, spatial, radiometric and Spectral resolutions. For optical sensor systems, imaging systems somehow offer a tradeoff between high spatial and high spectral resolution, and no single sensor offers both. Hence, in the remote sensing community, an image with 'greater quality' often means higher spatial or higher spectral resolution, which can only be obtained by more advanced sensors [1]. It is, therefore, necessary and very useful to be able to merge images with higher spectral information and higher spatial information [2].

Image fusion techniques can be classified into three categories depending on the stage at which fusion takes place; it is often divided into three levels, namely: pixel level, feature level and decision level of representation [3; 4]. The pixel image fusion techniques can be grouped into several techniques depending on the tools or the processing methods for image fusion procedure. By [5; 6] it is grouped into three classes: Color related techniques, statistical, arithmetic/numerical, and combined approaches.

The acronym IHS is sometimes permutated to HSI in the literature. IHS fusion methods are selected for comparison because they are the most widely used in commercial image processing systems. However, lots of papers reporting results of IHS sharpening technique and not those in which the formula of IHS transformation were used [7-19]. Many other papers describe different formula of IHS transformations, which have some important differences in the values of the matrix, are used such as IHS transformation [20-27].

The objectives of this study are to explain the different IHS transformations sharpening algorithms and experiment it as based image fusion techniques for remote sensing applications to fuse multispectral (MS) and Panchromatic (PAN) images. To remove that confusion, i. e IHS technique, this paper presents the most different formula of transformation matrix IHS it appears as well as effectiveness based image fusion and the performance of these methods. These are based on a comprehensive study that evaluates various PAN sharpening based on IHS techniques and programming in VB6 to achieve the fusion algorithm.

### I. IHS Fusion Technique

The IHS technique is one of the most commonly used fusion techniques for sharpening. It has become a standard procedure in image analysis for color enhancement, feature enhancement, improvement of spatial resolution and the fusion of disparate data sets [29]. In the IHS space, spectral information is mostly reflected on the hue and the saturation. From the visual system, one can conclude that the intensity change has little effect on the spectral information and is easy to deal with. For the fusion of the high-resolution and multispectral remote sensing images, the goal is ensuring the spectral information and adding the detail information of high spatial resolution, therefore, the fusion is even more adequate for treatment in IHS space [26].

Literature proposes many IHS transformation algorithms have been developed for converting the RGB

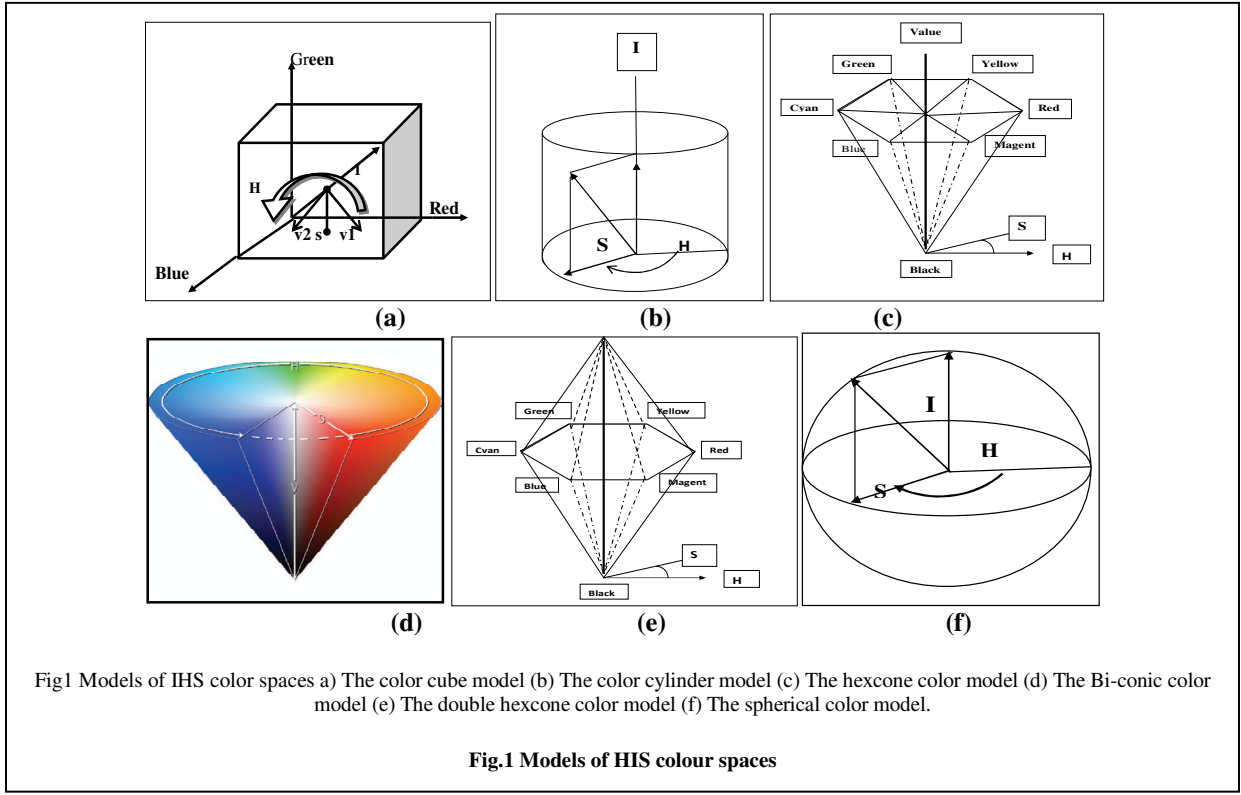

Fig1 Models of IHS color spaces a) The color cube model (b) The color cylinder model (c) The hexcone color model (d) The Bi-conic color model (e) The double hexcone color model (f) The spherical color model.

**Fig.1 Models of HIS colour spaces**

values. Some are also named HSV (hue, saturation, value) or HLS (hue, luminance, saturation). (Fig.1) illustrates the geometric interpretation. While the complexity of the models varies, they produce similar values for hue and saturation. However, the algorithms differ in the method used in calculating the intensity component of the transformation. The most common intensity definitions are [30]:

$$V = Max\{R, G, B\} \quad I = \frac{R+G+B}{3}$$
$$L = \frac{\max(R,G,B) + \min(R,G,B)}{2} \quad (1)$$

The first system (based on V), also known as the Smith's hexcone and the second system (based on L), known as Smith's triangle model [31]. The hexcone transformation of IHS is referred to as HSV model which drives its name from the parameters, hue, saturation, and value, the term "value" instead of "intensity" in this system.

Most literature recognizes IHS as a third-order method because it employs a 3×3 matrix as its transform kernel in the RGB IHS conversion model [32]. Many published studies show that various IHS transformations, which have some important differences in the values of the matrix, are used, which that description below. Here, this study denoted IHS1, IHS2, IHS3 …etc refer to the formula that used and R = Red, G = Green, B = Blue I = Intensity, H = Hue, S = Saturation, $v_1 v_2$ = Cartesian components of hue and saturation.

## A. HSV

The first IHS1 corresponding matrix expression of HSV is as follows [33]:

$$\begin{bmatrix} I \\ v_1 \\ v_2 \end{bmatrix} = \begin{bmatrix} 0.577 & 0.577 & 0.577 \\ -0.408 & -0.408 & 0.816 \\ -0.707 & 0.707 & 1.703 \end{bmatrix} \begin{bmatrix} R \\ G \\ B \end{bmatrix}$$
$$H = tan^{-1}\left(\frac{v2}{v1}\right), S = \sqrt{v_1^2 + v_2^2} \quad (2)$$

The gray value Pan image of a pixel is used as the value in the related color image, i.e. in the above equation (2) V=I [33]:

$$\begin{bmatrix} R \\ G \\ B \end{bmatrix} = \begin{bmatrix} 0.577 & -0.408 & -0.707 \\ 0.577 & -0.408 & 0.816 \\ 0.577 & 0.816 & 0 \end{bmatrix} \begin{bmatrix} I \\ v_1 \\ v_2 \end{bmatrix} \quad (3)$$

## B. IHS1:

One common IHS transformation is based on a cylinder color model, which is proposed by [34] and implemented in PCI Geomatica. The IHS coordinate system can be represented as a cylinder. The cylindrical transformation color model has the following equations:

$$\begin{bmatrix} I \\ v_1 \\ v_2 \end{bmatrix} = \begin{bmatrix} \frac{1}{\sqrt{3}} & \frac{1}{\sqrt{3}} & \frac{1}{\sqrt{3}} \\ \frac{-1}{\sqrt{6}} & \frac{-1}{\sqrt{6}} & \frac{2}{\sqrt{6}} \\ \frac{-1}{\sqrt{2}} & \frac{1}{\sqrt{2}} & 0 \end{bmatrix} \begin{bmatrix} R \\ G \\ B \end{bmatrix} \quad H = \tan^{-1}\left(\frac{v2}{v1}\right) \&$$

$$S = \sqrt{v_1^2 + v_2^2} \quad (4)$$

Where $v_1$ and $v_2$ are two intermediate values. In the algorithm there is a processing of special cases and a final scaling of the intensity, hue and saturation values between 0 and 255. The corresponding inverse transformation is defined as:

$$v_1 = S\cos(H) \& \ v_2 = S\sin(H) \quad (5)$$

$$\begin{bmatrix} R^{'} \\ G^{'} \\ B^{'} \end{bmatrix} = \begin{bmatrix} \frac{1}{\sqrt{3}} & \frac{-1}{\sqrt{6}} & \frac{-1}{\sqrt{2}} \\ \frac{1}{\sqrt{3}} & \frac{-1}{\sqrt{6}} & \frac{1}{\sqrt{2}} \\ \frac{1}{\sqrt{3}} & \frac{2}{\sqrt{6}} & 0 \end{bmatrix} \begin{bmatrix} I \\ v1 \\ v2 \end{bmatrix} \quad (6)$$

## C. IHS2:

Other color spaces that have simple computational transformations such as IHS coordinates within the RGB cube. The transformation is [35]:

$$\begin{bmatrix} I \\ v_1 \\ v_2 \end{bmatrix} = \begin{bmatrix} \frac{1}{3} & \frac{1}{3} & \frac{1}{3} \\ \frac{-1}{\sqrt{6}} & \frac{-1}{\sqrt{6}} & \frac{2}{\sqrt{6}} \\ \frac{1}{\sqrt{6}} & \frac{-2}{\sqrt{6}} & 0 \end{bmatrix} \begin{bmatrix} R \\ G \\ B \end{bmatrix}$$

$$H = \tan^{-1}\left(\frac{v2}{v1}\right) \text{ in the rang 0 to 360 } \&$$

$$S = \sqrt{v_1^2 + v_2^2} \quad (7)$$

The corresponding inverse transformation this is given by [35]:

$$\begin{bmatrix} R^{'} \\ G^{'} \\ B^{'} \end{bmatrix} = \begin{bmatrix} 1 & -0.204124 & 0.612372 \\ 1 & -0.204124 & -0.612372 \\ 1 & 0.408248 & 0 \end{bmatrix} \begin{bmatrix} I \\ v_1 \\ v_2 \end{bmatrix}$$

$$v_1 = S\cos(2\pi H) \& \ v_2 = S\sin(2\pi H) \quad (8)$$

## D. IHS3:

[24] is one of these studies. The transformation model for IHS transformation is the one below:

$$\begin{bmatrix} I \\ v_1 \\ v_2 \end{bmatrix} = \begin{bmatrix} \frac{1}{3} & \frac{1}{3} & \frac{1}{3} \\ \frac{-1}{\sqrt{6}} & \frac{-1}{\sqrt{6}} & \frac{2}{\sqrt{6}} \\ \frac{1}{\sqrt{6}} & \frac{-1}{\sqrt{6}} & 0 \end{bmatrix} \begin{bmatrix} R \\ G \\ B \end{bmatrix} \quad H = \tan^{-1}\left(\frac{v2}{v1}\right) \&$$

$$S = \sqrt{v_1^2 + v_2^2} \quad (9)$$

$$\begin{bmatrix} R^{'} \\ G^{'} \\ B^{'} \end{bmatrix} = \begin{bmatrix} 1 & \frac{-1}{\sqrt{6}} & \frac{3}{\sqrt{6}} \\ 1 & \frac{-1}{\sqrt{6}} & \frac{-3}{\sqrt{6}} \\ 1 & \frac{2}{\sqrt{6}} & 0 \end{bmatrix} \begin{bmatrix} I \\ v_1 \\ v_2 \end{bmatrix} \quad (10)$$

## E. IHS4:

[29] propose an IHS transformation taken from Harrison and Jupp (1990).

$$\begin{bmatrix} I \\ v_1 \\ v_2 \end{bmatrix} = \begin{bmatrix} \frac{1}{3} & \frac{1}{3} & \frac{1}{3} \\ \frac{1}{\sqrt{6}} & \frac{1}{\sqrt{6}} & \frac{-2}{\sqrt{6}} \\ \frac{1}{\sqrt{2}} & \frac{-1}{\sqrt{2}} & 0 \end{bmatrix} \begin{bmatrix} R \\ G \\ B \end{bmatrix} \quad H = \tan^{-1}\left(\frac{v1}{v2}\right)$$

$$\& \ S = \sqrt{v_1^2 + v_2^2} \quad (11)$$

The corresponding inverse transformation is defined as:

$$\begin{bmatrix} R^{'} \\ G^{'} \\ B^{'} \end{bmatrix} = \begin{bmatrix} \frac{1}{\sqrt{3}} & \frac{1}{\sqrt{6}} & \frac{1}{\sqrt{2}} \\ \frac{1}{\sqrt{3}} & \frac{1}{\sqrt{6}} & \frac{-1}{\sqrt{2}} \\ \frac{1}{\sqrt{3}} & \frac{-2}{\sqrt{6}} & 0 \end{bmatrix} \begin{bmatrix} I \\ v1 \\ v2 \end{bmatrix} \quad (12)$$

## F. IHS5:

[22] Proposes an IHS transformation taken from Carper et al. 1990. The transformation is:

$$\begin{bmatrix} I \\ v_1 \\ v_2 \end{bmatrix} = \begin{bmatrix} \frac{1}{3} & \frac{1}{3} & \frac{1}{3} \\ \frac{1}{\sqrt{6}} & \frac{1}{\sqrt{6}} & \frac{-2}{\sqrt{6}} \\ \frac{1}{\sqrt{2}} & \frac{-1}{\sqrt{2}} & 0 \end{bmatrix} \begin{bmatrix} R \\ G \\ B \end{bmatrix} \quad H = \tan^{-1}\left(\frac{v2}{v1}\right) \&$$

$$S = \sqrt{v_1^2 + v_2^2} \quad (13)$$

$$\begin{bmatrix} R^{'} \\ G^{'} \\ B^{'} \end{bmatrix} = \begin{bmatrix} 1 & \frac{1}{\sqrt{6}} & \frac{1}{\sqrt{2}} \\ 1 & \frac{1}{\sqrt{6}} & \frac{-1}{2} \\ 1 & \frac{-2}{\sqrt{6}} & 0 \end{bmatrix} \begin{bmatrix} I \\ v_1 \\ v_2 \end{bmatrix} \quad (14)$$

## G. IHS6:

[28] Used the linear transformation model for IHS transformation is the one below as well as the IHS transformation taken from Carper et.al. 1990, but different matrix compared to [22]. [28] Published an article on IHS-like image fusion methods. The transformation is:

$$\begin{bmatrix} I_V \\ v_1 \\ v_2 \end{bmatrix} = \begin{bmatrix} \frac{1}{3} & \frac{1}{3} & \frac{1}{3} \\ \frac{\sqrt{2}}{6} & \frac{\sqrt{2}}{6} & \frac{\sqrt{2}}{6} \\ \frac{1}{\sqrt{2}} & \frac{-1}{\sqrt{2}} & 0 \end{bmatrix} \begin{bmatrix} R \\ G \\ B \end{bmatrix} \quad H = \tan^{-1}\left(\frac{v_2}{v_1}\right) \&$$

$$S = \sqrt{v_1^2 + v_2^2} \quad (15)$$

$$\begin{bmatrix} R^{'} \\ G^{'} \\ B^{'} \end{bmatrix} = \begin{bmatrix} 1 & \frac{-1}{\sqrt{2}} & \frac{1}{\sqrt{2}} \\ 1 & \frac{-1}{\sqrt{2}} & \frac{-1}{\sqrt{2}} \\ 1 & \sqrt{2} & 0 \end{bmatrix} \begin{bmatrix} I_P \\ v_1 \\ v_2 \end{bmatrix} \quad (16)$$

They modified the matrix by using parameters. The following formula modified result:

$$\begin{bmatrix} R' \\ G' \\ B' \end{bmatrix} = \begin{bmatrix} 1 & \frac{-1}{\sqrt{2}} & \frac{1}{\sqrt{2}} \\ 1 & \frac{-1}{\sqrt{2}} & \frac{-1}{\sqrt{2}} \\ 1 & \sqrt{2} & 0 \end{bmatrix} \begin{bmatrix} \alpha I_P + \beta I_{MS_k} \\ v_1 \\ v_2 \end{bmatrix} \quad (17)$$

Where $\alpha, \beta$ are the fused parameters, which $0 \le \alpha$ ,$\beta \le 1$ and $I_P I_{MS_k}$ ; the intensity is of each P and $MS_k$ image respectively.

### H. HLS:

[26] propose that HLS transformation is alternative to IHS. The transformation from RGB to LHS color space is the same propose [22]. But the transformation back to RGB space gives different results the transformation is:

$$\begin{bmatrix} I \\ v_1 \\ v_2 \end{bmatrix} = \begin{bmatrix} \frac{1}{3} & \frac{1}{3} & \frac{1}{3} \\ \frac{1}{\sqrt{6}} & \frac{1}{\sqrt{6}} & \frac{-2}{\sqrt{6}} \\ \frac{1}{\sqrt{2}} & \frac{-1}{\sqrt{2}} & 0 \end{bmatrix} \begin{bmatrix} R \\ G \\ B \end{bmatrix} \quad H = \tan^{-1}\left(\frac{v1}{v2}\right) \; \&$$

$$S = \sqrt{v_1^2 + v_2^2} \quad (18)$$

$$\begin{bmatrix} R' \\ G' \\ B' \end{bmatrix} = \begin{bmatrix} 1 & \frac{1}{\sqrt{6}} & \frac{1}{\sqrt{2}} \\ 1 & \frac{1}{\sqrt{6}} & \frac{-1}{\sqrt{2}} \\ 1 & \frac{-2}{\sqrt{6}} & 0 \end{bmatrix} \begin{bmatrix} I \\ v_1 \\ v_2 \end{bmatrix} \quad (19)$$

### I. IHS7:

[36] Published an article on modified IHS –like image fusion methods. He uses the basic equations of the cylindrical model to convert from RGB space to IHS space and back into RGB space are below as well as Modified it for more information refer to [36]. The transformation is:

$$\begin{bmatrix} I \\ v_1 \\ v_2 \end{bmatrix} = \begin{bmatrix} \frac{1}{3} & \frac{1}{3} & \frac{1}{3} \\ \frac{1}{2} & \frac{-1}{2} & 1 \\ \frac{\sqrt{3}}{2} & \frac{-\sqrt{3}}{2} & 0 \end{bmatrix} \begin{bmatrix} R \\ G \\ B \end{bmatrix} \quad (20)$$

$$H = \begin{cases} 0 & \text{if } v1 = 0 \text{ and } v2 = 0 \\ \tan^{-1}\left(\frac{v2}{v1}\right) + 2\pi & \text{if } v1 \ge 0 \text{ and } v2 < 0 \\ \tan^{-1}\left(\frac{v2}{v1}\right) & \text{if } v1 \ge 0 \text{ and } v2 \ge 0 \\ \tan^{-1}\left(\frac{v2}{v1}\right) + \pi & \text{if } v1 < 0 \end{cases} \; \&$$

$$S = \sqrt{v_1^2 + v_2^2} \quad (21)$$

$$\begin{bmatrix} R' \\ G' \\ B' \end{bmatrix} = \begin{bmatrix} 1 & \frac{-1}{3} & \frac{1}{\sqrt{3}} \\ 1 & \frac{-1}{3} & \frac{-1}{\sqrt{3}} \\ 1 & \frac{2}{3} & 0 \end{bmatrix} \begin{bmatrix} I \\ v_1 \\ v_2 \end{bmatrix} \quad v_1 = S\cos(H) \; \&$$

$$v_2 = S \sin(H) \quad (22)$$

### J. YIQ:

Another color encoding system called YIQ (Fig.2) has a straightforward transformation from RGB with no loss of information. The YIQ model was designed to take advantage of the human system's greater sensitivity to changes in luminance than to changes in hue or saturation [37]. In the case of the YIQ transformations, the component Y represents the luminance of a color, while its chrominance is denoted by I and Q signals [38].Y is just the brightness of a panchromatic monochrome image. It combines the red, green, and blue signals in proportion to the human eye's sensitivity to them .The I signal is essentially red minus cyan, while Q is magenta minus green [39] express jointly hue and saturation . The relationship between YIQ and RGB is given as follows: [30; 37]:

$$\begin{bmatrix} Y \\ I \\ Q \end{bmatrix} = \begin{bmatrix} 0.299 & 0.587 & 0.144 \\ 0.596 & -0.274 & 0.322 \\ 0.211 & -0.523 & 0.312 \end{bmatrix} \begin{bmatrix} R \\ G \\ B \end{bmatrix} \quad (23)$$

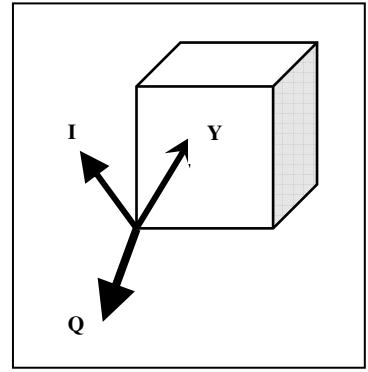

Fig.2: Geometric Relations in RGB - YIQ Model

The YIQ is transformed inversely to RGB space is given by [30; 37]:

$$\begin{bmatrix} R' \\ G' \\ B' \end{bmatrix} = \begin{bmatrix} 1 & 0.956 & 0.621 \\ 1 & -0.272 & -0.647 \\ 1 & -1.106 & 1.703 \end{bmatrix} \begin{bmatrix} Y \\ I \\ Q \end{bmatrix} \quad (24)$$

## II. The IHS-Based PAN Sharpening

The IHS pan sharpening technique is the oldest known data fusion method and one of the simplest. Fig.3 illustrates this technique for convenience. In this technique the following steps are performed:

1. The low resolution MS imagery is co-registered to the same area as the high resolution PAN imagery and resampled to the same resolution as the PAN imagery.
2. The three resampled bands of the MS imagery, which represent the RGB space, are transformed into IHS components.
3. The PAN imagery is histogram matched to the 'I' component. This is done in order to compensate for the spectral differences

between the two images, which occurred due to different sensors or different acquisition dates and angles.

4. The intensity component of MS imagery is replaced by the histogram matched PAN imagery. The RGB of the new merged MS imagery is obtained by computing a reverse IHS to RGB transform.

To evaluate the ability of enhancing spatial details and preserving spectral information, some Indices including Standard Deviation (SD), Entropy (En), Correlation Coefficient (CC), Signal-to Noise Ratio (SNR), Normalization Root Mean Square Error (NRMSE) and Deviation Index (DI) of the image these measures given in (Table 1), and the results are shown in Table 2. In the

| | Equation |
|---|---|
| CC | $CC = \dfrac{\sum_i^n \sum_j^m (F_k(i,j) - \overline{F}_k)(M_k(i,j) - \overline{M}_k)}{\sqrt{\sum_i^n \sum_j^m (F_k(i,j) - \overline{F}_k)^2}\sqrt{\sum_i^n \sum_j^m (M_k(i,j) - \overline{M}_k)^2}}$ |
| En | $En = -\sum_0^{I-1} P(i)\log_2 P(i)$ |
| DI | $DI_k = \dfrac{1}{nm}\sum_i^n \sum_j^m \dfrac{\lvert F_k(i,j) - M_k(i,j)\rvert}{M_k(i,j)}$ |
| SNR | $SNR_k = \sqrt{\dfrac{\sum_i^n \sum_j^m (F_k(i,j))^2}{\sum_i^n \sum_j^m (F_k(i,j) - M_k(i,j))^2}}$ |
| NRM SE | $NRMSE_k = \sqrt{\dfrac{1}{nm*255^2}\sum_i^n \sum_j^m (F_k(i,j) - M_k(i,j))^2}$ |

## III. RESULTS AND DISCUSSION

In order to validate the theoretical analysis, the performance of the methods discussed above was further evaluated by experimentation. Data sets used for this study were collected by the Indian IRS-1C PAN (0.50 - 0.75 µm) of the (5.8 $m$) resolution panchromatic band. Where the American Landsat (TM) the red (0.63 - 0.69 µm), green (0.52 - 0.60 µm) and blue (0.45 - 0.52 µm) bands of the 30 m resolution multispectral image were used in this experiment. Fig. 4a.&b shows the IRS-1C PAN and multispectral TM images. The scenes covered the same area of the Mausoleums of the Chinese Tang – Dynasty in the PR China [40] was selected as test sit in this study. Since this study is involved in evaluation of the effect of the various spatial, radiometric and spectral resolution for image fusion, an area contains both manmade and natural features is essential to study these effects. Hence, this work is an attempt to study the quality of the images fused from different sensors with various characteristics. The size of the PAN is 600 * 525 pixels at 6 bits per pixel and the size of the original multispectral is 120 * 105 pixels at 8 bits per pixel, but this is upsampled to by Nearest neighbor was used to avoid spectral contamination caused by interpolation. The pairs of images were geometrically registered to each other.

The Fig. 5 shows quantitative measures for the fused images for the various fusion methods. It can be seen that the standard deviation of the fused images remain constant for all methods except HSV, IHS6 and IHS7.

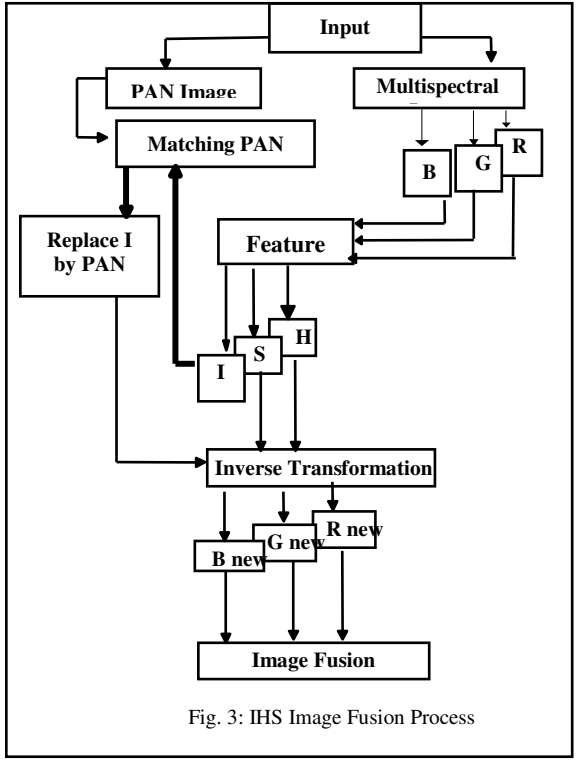

Fig. 3: IHS Image Fusion Process

following sections, $F_k$, $M_k$ are the measurements of each the brightness values of homogenous pixels of the result image and the original multispectral image of band k, $\overline{M}_k$ and $\overline{F}_k$ are the mean brightness values of both images and are of size $n * m$. $BV$ is the brightness value of image data $\overline{M}_k$ and $\overline{F}_k$.

Table 1: Indices Used to Assess Fusion Images.

| Item | Equation |
|---|---|
| SD | $\sigma = \sqrt{\dfrac{\sum_{i=1}^m \sum_{j=1}^n (BV(n,m) - \mu)^2}{m \times n}}$ |

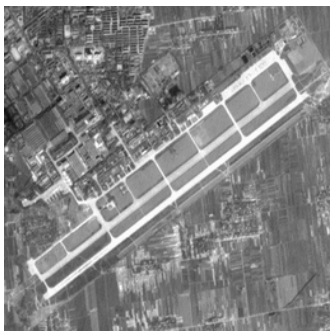

Fig. 4a. Original panchromatic

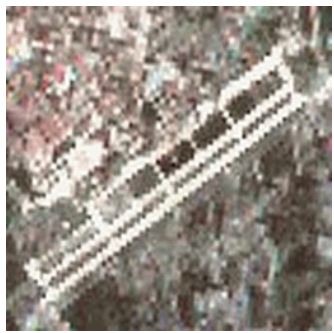

Fig. 4b. Original Multispectral

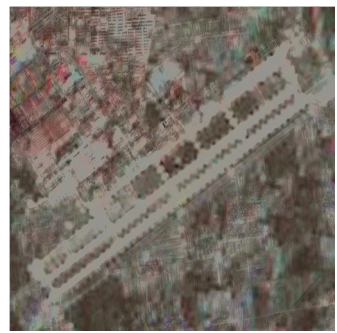

Fig.4c.HSV

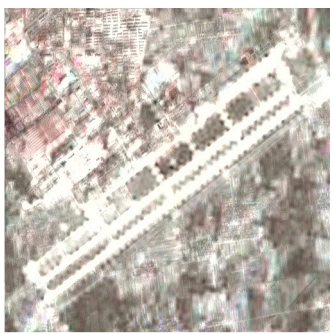

Fig.4d.IHS1

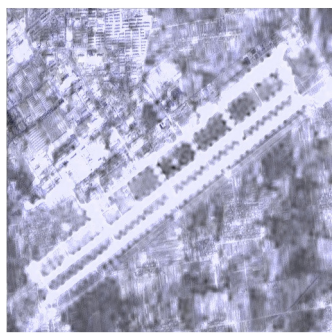

Fig.4e.IHS2

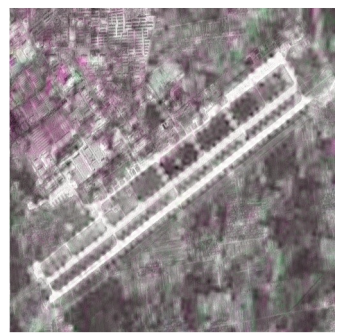

Fig.4f.IHS3

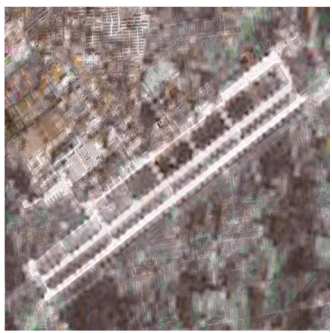

Fig.4g. IHS4

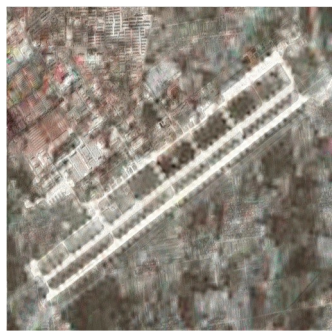

Fig.4h. IHS5

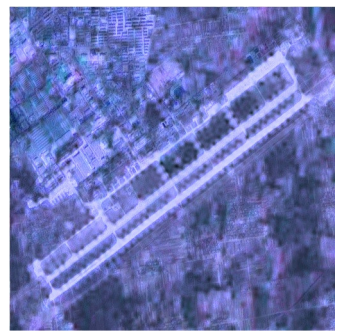

Fig.4i. IHS6

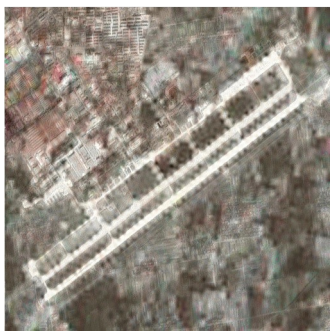

Fig.4j.HLS

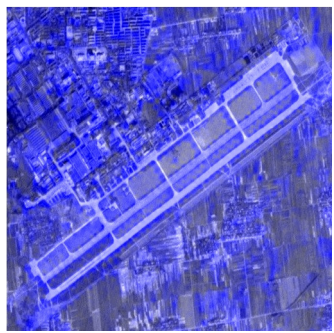

Fig.4k.IHS7

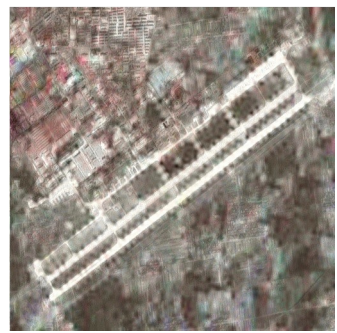

Fig.4l. YIQ

Fig(4) Original and Fused images

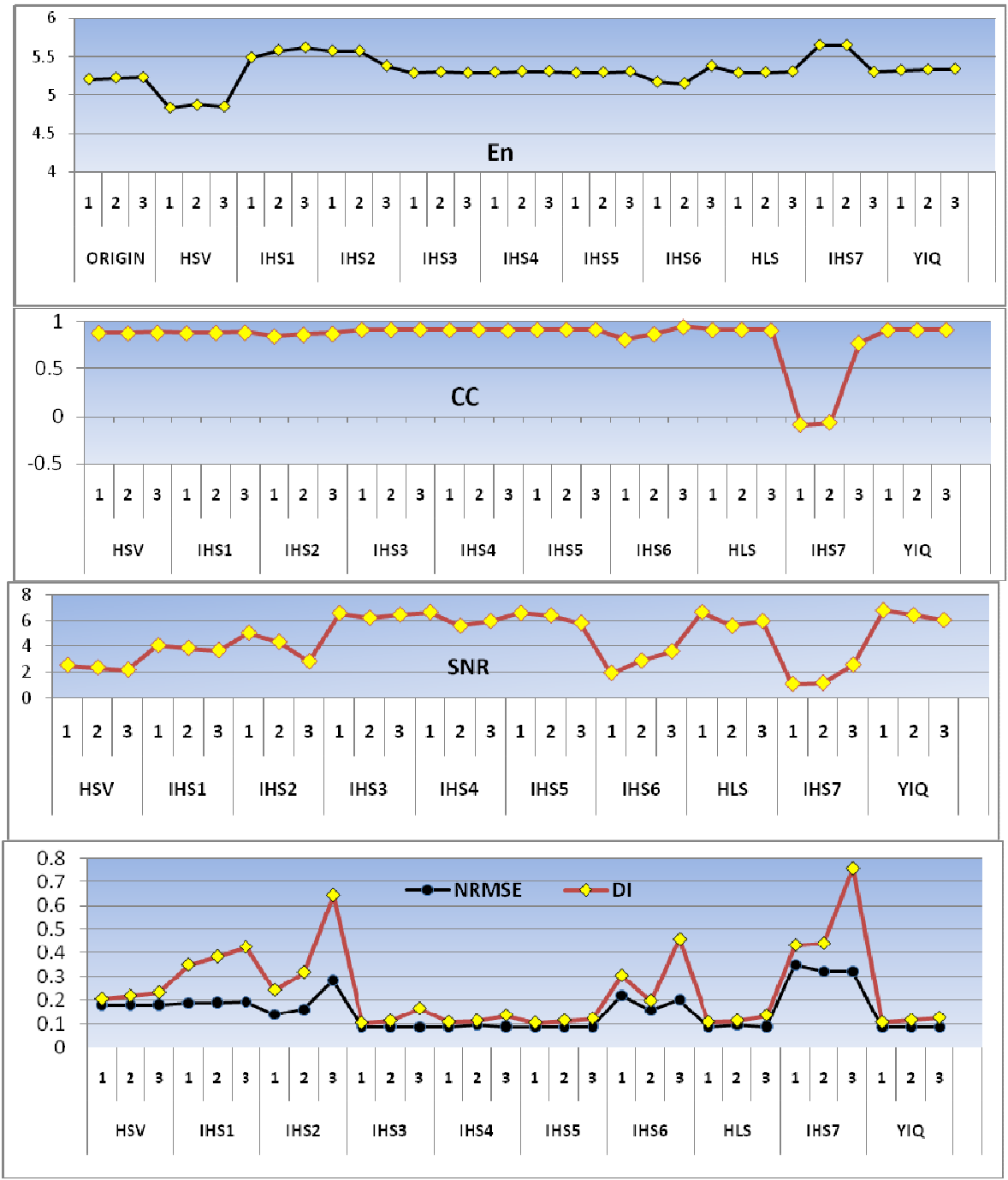

Fig. 5: Chart Representation of En, CC, SNR, NRMSE & DI of Fused Image

Table 2: Quantitative Analysis of Origional MS and Fused Image Results

| Method | Band | SD | En | SNR | NRMSE | DI | CC |
|--------|------|------|------|------|-------|------|------|
| ORIGIN | Red | 51.02 | 5.2093 | | | | |
| | Green | 51.48 | 5.2263 | | | | |
| | Blue | 51.98 | 5.2326 | | | | |
| HSV | Red | 25.91 | 4.8379 | 2.529 | 0.182 | 0.205 | 0.881 |
| | Green | 26.822 | 4.8748 | 2.345 | 0.182 | 0.218 | 0.878 |
| | Blue | 27.165 | 4.8536 | 2.162 | 0.182 | 0.232 | 0.883 |
| IHS1 | Red | 43.263 | 5.4889 | 4.068 | 0.189 | 0.35 | 0.878 |
| | Green | 45.636 | 5.5822 | 3.865 | 0.191 | 0.384 | 0.882 |
| | Blue | 46.326 | 5.6178 | 3.686 | 0.192 | 0.425 | 0.885 |
| IHS2 | Red | 41.78 | 5.5736 | 5.038 | 0.138 | 0.242 | 0.846 |
| | Green | 41.78 | 5.5736 | 4.337 | 0.16 | 0.319 | 0.862 |
| | Blue | 44.314 | 5.3802 | 2.82 | 0.285 | 0.644 | 0.872 |
| IHS3 | Red | 41.13 | 5.2877 | 6.577 | 0.088 | 0.103 | 0.915 |
| | Green | 42.32 | 5.3015 | 6.208 | 0.088 | 0.112 | 0.915 |
| | Blue | 41.446 | 5.2897 | 6.456 | 0.086 | 0.165 | 0.917 |
| IHS4 | Red | 41.173 | 5.2992 | 6.658 | 0.087 | 0.107 | 0.913 |
| | Green | 42.205 | 5.3098 | 5.593 | 0.095 | 0.113 | 0.915 |
| | Blue | 42.889 | 5.3122 | 5.954 | 0.088 | 0.136 | 0.908 |
| IHS5 | Red | 41.164 | 5.291 | 6.583 | 0.088 | 0.104 | 0.915 |
| | Green | 41.986 | 5.2984 | 6.4 | 0.086 | 0.114 | 0.917 |
| | Blue | 42.709 | 5.3074 | 5.811 | 0.088 | 0.122 | 0.917 |
| IHS6 | Red | 35.664 | 5.172 | 1.921 | 0.221 | 0.304 | 0.811 |
| | Green | 33.867 | 5.1532 | 2.881 | 0.158 | 0.197 | 0.869 |
| | Blue | 47.433 | 5.3796 | 3.607 | 0.203 | 0.458 | 0.946 |
| HLS | Red | 41.173 | 5.291 | 6.657 | 0.087 | 0.107 | 0.913 |
| | Green | 42.206 | 5.2984 | 5.592 | 0.095 | 0.113 | 0.915 |
| | Blue | 42.889 | 5.3074 | 5.954 | 0.088 | 0.136 | 0.908 |
| IHS7 | Red | 35.121 | 5.6481 | 1.063 | 0.35 | 0.433 | -0.087 |
| | Green | 35.121 | 5.6481 | 1.143 | 0.325 | 0.44 | -0.064 |
| | Blue | 37.78 | 5.3008 | 2.557 | 0.323 | 0.758 | 0.772 |
| YIQ | Red | 41.691 | 5.3244 | 6.791 | 0.086 | 0.106 | 0.912 |
| | Green | 42.893 | 5.3334 | 6.415 | 0.086 | 0.115 | 0.912 |
| | Blue | 43.359 | 5.3415 | 6.035 | 0.086 | 0.125 | 0.914 |

Correlation values also remain practically constant, very near the maximum possible value except IHS6 and HS7. The differences between the reference image and the fused images during $SD$ & $CC$ values are so small that they do not bear any real significance. This is due to the fact that, the Matching processing of the intensity of MS and PAN images by mean and standard division was done before the merging processing. But with the results of $SNR$, $NRMSE$ and $DI$ appear changing significantly. It can be observed that from the diagram of Fig. 5. That the fused image the results of $NRMSE$ & $DI$ show that the IHS5 and YIQ methods give best results with respect to the other methods followed by the HLS and IHS4who get the same values presented the lowest value of the $NRMSE$ & $DI$ as well as the higher of the $SNR$. Hence, the spectral qualities of fused images by the IHS5 and YIQ methods are much better than the others. In contrast, It can also be noted that the IHS7, HS6, IHS2, IHS1 images produce highly $NRMSE$ & $DI$ values indicate that these methods deteriorate spectral information content for the reference image.

## IV. CONCLUSION

In this study the different formulas of transformation matrix IHS it as well as the effectiveness of the based image fusion and the performance of these methods. The IHS transformations based fusion show different results corresponding to the formula of IHS transformation that is used. In a comparison to spatial effects, it can be seen that the results of the four formulas of IHS transformation methods by IHS5, YIQ, HLS and IHS4 display the same details. But the statistical analysis of the different formulas of IHS transformation based fusion show that the spectral effect by IHS5 and YIQ methods presented here are the best of the methods studied. The use of the formula of IHS transformation based fusion methods by IHS5 and YIQ could, therefore, be strongly recommended if the goal of the merging is to achieve the best representation of the spectral information of multispectral image and the spatial details of a high-resolution panchromatic image.

The results that can be reported here are: 1- some of the statistical evaluation methods do not bear any real significance such as $SD$, $En$ and $CC$. 2- The analytical technique of DI is much more useful for measuring the spectral distortion than NRMSE. 3-Since the NRMSE gave the same results for some methods, but the DI gave the smallest different ratio between those methods, therefore , it is strongly recommended to use the $DI$ because of its mathematical more precision as quality indicator.

## AKNOWLEDGEMENTS

The Authors wish to thank Dr. Fatema Al-Kamissi at University of Ammran (Yemen) for her suggestion and comments. The authors would also like to thank the anonymous reviewers for their helpful comments and suggestions.

**AUTHORS**

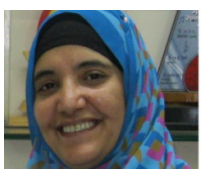

*Firouz Abdullah Al-Wassai*. Received the B.Sc. degree in, Physics from University of Sana'a, Yemen, Sana'a, in 1993. The M.Sc.degree in, Physics from Bagdad University , Iraqe, in 2003, Research student.Ph.D in the department of computer science (S.R.T.M.U), India, Nanded.

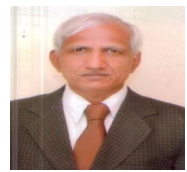

*Dr. N.V. Kalyankar*, Principal,Yeshwant Mahvidyalaya, Nanded(India) completed M.Sc.(Physics) from Dr. B.A.M.U, Aurangabad. In 1980 he joined as a leturer in department of physics at Yeshwant Mahavidyalaya, Nanded. In 1984 he completed his DHE. He completed his Ph.D. from Dr.B.A.M.U. Aurangabad in 1995. From 2003 he is working as a Principal to till date in Yeshwant Mahavidyalaya, Nanded. He is also research guide for Physics and Computer Science in S.R.T.M.U, Nanded. 03 research students are successfully awarded Ph.D in Computer Science under his guidance. 12 research students are successfully awarded M.Phil in Computer Science under his guidance He is also worked on various boides in S.R.T.M.U, Nanded. He is also worked on various bodies is S.R.T.M.U, Nanded. He also published 30 research papers in various international/national journals. He is peer team member of NAAC (National Assessment and Accreditation Council, India ). He published a book entilteld "DBMS concepts and programming in Foxpro". He also get various educational wards in which "Best Principal" award from S.R.T.M.U, Nanded in 2009 and "Best Teacher" award from Govt. of Maharashtra, India in 2010. He is life member of Indian "Fellowship of Linnean Society of London(F.L.S.)" on 11 National Congress, Kolkata (India). He is also honored with November 2009.

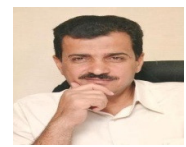

*Dr. Ali A. Al-Zuky*. B.Sc Physics Mustansiriyah University, Baghdad , Iraq, 1990. M Sc. In1993 and Ph. D. in1998 from University of Baghdad, Iraq. He was supervision for 40 postgraduate students (MSc. & Ph.D.) in different fields (physics, computers and Computer Engineering and Medical Physics). He has More than 60 scientific papers published in scientific journals in several scientific conferences.